# Potential sources of dataset bias complicate investigation of underdiagnosis by machine learning algorithms

Mélanie Bernhardt, Charles Jones, Ben Glocker

Department of Computing, Imperial College London, London, SW72AZ, United Kingdom

The authors contributed equally to this work. Correspondence: b.glocker@imperial.ac.uk

An increasing number of reports raise concerns about the risk that machine learning algorithms could amplify health disparities due to biases embedded in the training data[1,2]. Seyyed-Kalantari et al.[3] find that models trained on three chest X-ray datasets yield disparities in false-positive rates (FPR) across subgroups on the 'no-finding' label (indicating the absence of disease). The models consistently yield higher FPR on subgroups known to be historically underserved, and the study concludes that the models exhibit and potentially even amplify systematic underdiagnosis. We argue that the experimental setup in the study is insufficient to study algorithmic underdiagnosis. In the absence of specific knowledge (or assumptions) about the extent and nature of the dataset bias, it is difficult to investigate model bias. Importantly, their use of test data exhibiting the same bias as the training data (due to random splitting) severely complicates the interpretation of the reported disparities.

The datasets used in the study by Seyyed-Kalantari et al. may suffer from multiple sources of bias. Based on the data characteristics and study assumptions, it seems plausible to consider at least three types of bias causing dataset shift across patient subgroups[4,5]: (i) population shift (the majority of subjects is White and there are significant differences in age); (ii) prevalence shift (large variation in the presence of different disease conditions); and (iii) annotation shift (as a result of assumed physician underdiagnosis where specific subgroups may be systematically mislabelled more often than others). It is important to consider how these different dataset biases affect AI models if we want to understand the observed disparities across subgroups.

Under population and prevalence shift, performance disparities arise from imbalance and mismatch of subgroup characteristics (such as change in age or disease distribution)[6], resulting in jointly shifted true and false-positive rates (TPR/FPR) similar to the ones reported in the underdiagnosis study. However, under these shifts the underlying mapping from imaging features to disease labels, is consistent and may remain valid across subgroups. For that reason, these disparities can be corrected by incorporating prior knowledge about the (expected) ideal population using re-sampling techniques[7], re-weighting of the training objective, or calibration of decision thresholds[8]. Seyyed-Kalantari et al. discuss the potential flaws of group-specific threshold selection, and instead use a single threshold optimised over the whole patient population. However, given that subgroups will differ substantially due to population and prevalence shift, a single threshold is expected to yield disparate performance. For CheXpert, the presence of 'no-finding' is 7%, 9%, 9% in females over the age of 40 identifying as White, Asian, and Black. This increases to 15%, 14%, and 21% in females under the age of 40. These mixed effects of population and prevalence shift will contribute to the increase in FPR which was predominantly observed in young patients.

Underdiagnosis, on the other hand, is a severe form of systematic mislabelling causing annotation shift that is much more difficult to handle. A model trained under annotation shift cannot be expected to perform well across subgroups, as the mapping from imaging features to disease labels is inconsistent. Similar patterns that are labelled as disease in one group may be labelled as 'no-finding' in a group suffering from underdiagnosis. This cannot be corrected for, and the most effective (if not only) mitigation strategy would be to re-annotate the data to obtain unbiased disease labels[9]. These fundamental differences highlight the importance to distinguish between population, prevalence, and annotation shift as potential causes of algorithmic bias. Defining algorithmic underdiagnosis simply as an increase in FPR for 'no-finding', without the ability to attribute an underlying cause, severely limits the conclusions one may draw and misses the opportunity for identifying possible mitigation.

One may argue that the exact reason causing the disparities may clinically not be relevant as the consequences for patients could be equally detrimental. In the presence of dataset bias, however, even assessing whether a model is biased or not is difficult. In the case of an unfair model that has picked up bias from the training data, one can expect that this bias is replicated at test-time. However, given that the models in the underdiagnosis study are evaluated on test data exhibiting the same bias as the training data (due to random splitting of the original datasets), we may actually not observe any disparities for an unfair model. Conversely, a fair model would be expected to produce disparities across subgroups at test-time because it correctly classifies mislabelled images in the group suffering from underdiagnosis. Hence, in order to associate observed disparities with algorithmic underdiagnosis, one would need to assume that the test set is unbiased (meaning the diagnostic labels are largely correct in all subgroups). However, this assumption would then also apply to the training data as it is drawn from the same data distribution. This seems contradictory in the setting where algorithmic bias is picked up from biased training data.

A further difficulty stems from the fact that the used datasets were annotated using natural language processing methods generating labels based on radiology reports instead of expert labels or confirmed clinical outcomes. A subset of the CheXpert disease labels has been benchmarked against radiologists' annotations, showing an F1-score as low as 0.76 for 'no-finding'[10]. This suggests the presence of a substantial amount of label noise affecting the reliability of the reported results, in particular when the effect size of interest is not known. Regardless of underlying biases, label noise is of particular concern, given the small sample sizes for some of the intersectional subgroups.

Our discussion illustrates that investigating algorithmic underdiagnosis on biased and noisy datasets is difficult. While we do not suggest that the models inspected by Seyyed-Kalantari et al. are fair, the experimental setting used to study algorithmic fairness may not be suitable as algorithmic bias cannot be decoupled from dataset bias. It has been argued previously that causal assumptions about the data generation process may need to be incorporated[4]. Dataset curation, theoretical analyses, and simulation experiments will be key components of future studies on algorithmic bias in medical imaging and beyond.

## Acknowledgments

We would like to thank Xiaoxuan Liu, Alastair Denniston, and Melissa McCradden for the helpful discussions.

## Funding sources

M.B. is funded through an Imperial College London President's PhD Scholarship. C.J. is supported by Microsoft Research and EPSRC through the Microsoft PhD Scholarship Programme. B.G. is supported through funding from the European Research Council (ERC) under the European Union's Horizon 2020 research and innovation programme (Grant Agreement No. 757173, Project MIRA, ERC-2017-STG).

## Author contributions

The authors contributed equally to this work in terms of formulating the arguments, interpreting the available evidence, and co-writing the manuscript.

## Competing interests

B.G. is part-time employee of HeartFlow and Kheiron Medical Technologies and holds stock options with both as part of the standard compensation package. M.B. and C.J. declare no competing interests related to the manuscript.